\documentclass[letterpaper, 10 pt, conference]{ieeeconf}  

\IEEEoverridecommandlockouts                              

\usepackage{todonotes} 
\usepackage{balance}
\usepackage[fleqn]{amsmath}
\usepackage{algorithm}
\usepackage[noEnd = true]{algpseudocodex}
\usepackage{acronym}
\usepackage{mathtools}
\usepackage[export]{adjustbox}
\usepackage{siunitx}
\usepackage{booktabs}

\makeatletter
\let\NAT@parse\undefined
\makeatother
\usepackage[bookmarks=true, colorlinks=true, citecolor=blue, linkcolor=blue]{hyperref}
\usepackage[capitalise]{cleveref}

\acrodef{CNN}{Convolutional Neural Network}
\acrodef{CoM}{Center of Mass}
\acrodef{DoF}{Degree of Freedom}
\acrodef{FF}{Feed Forward}
\acrodef{IK}{Inverse Kinematics}
\acrodef{IMU}{Inertial Measurement Unit}
\acrodef{KF}{Kalman Filter}
\acrodef{LUT}{Look-Up Table}
\acrodef{ML}{Machine Learning}
\acrodef{MLP}{Multilayer Perceptron}
\acrodef{NN}{Neural Network}
\acrodef{PPO}{Proximal Policy Optimization}
\acrodef{RL}{Reinforcement Learning}
\acrodef{ROS}{Robot Operating System}
\acrodef{TCN}{Temporal Convolution Network}
\acrodef{THO}{Task Hierarchical Optimization}
\acrodef{MAE}{Mean Absolute Error}

\title{\LARGE \bf Reinforcement Learning Control for Autonomous Hydraulic Material Handling Machines with Underactuated Tools}
\author{Filippo A. Spinelli$^{1}$, Pascal Egli$^{1}$, Julian Nubert$^{1,2}$, Fang Nan$^{1}$, Thilo Bleumer$^{3}$, Patrick Goegler$^{3}$, \\ Stephan Brockes$^{3}$, Ferdinand Hofmann$^{3}$, and Marco Hutter$^{1}$%
\thanks{*This work is supported in part by the NCCR digital fabrication and robotics, the Liebherr-Hydraulikbagger GmbH, and the Max Planck ETH Center for Learning Systems.}
\thanks{$^{1}$The authors are with the Robotic Systems Lab, ETH Z\"urich, Z\"{u}rich, Switzerland.}
\thanks{$^{2}$The author is with the MPI for Intelligent Systems, Stuttgart, Germany.}
\thanks{$^{3}$The authors are with the Liebherr-Hydraulikbagger GmbH, Kirchdorf an der Iller, Germany.}
\thanks{Corresponding author: Filippo A. Spinelli, \href{mailto:fspinelli@ethz.ch}{\texttt{fspinelli@ethz.ch}}}
}

\begin{document}

\maketitle
\thispagestyle{empty}
\pagestyle{empty}

\begin{abstract}
The precise and safe control of heavy material handling machines presents numerous challenges due to the hard-to-model hydraulically actuated joints and the need for collision-free trajectory planning with a free-swinging end-effector tool. 
In this work, we propose an RL-based controller that commands the cabin joint and the arm simultaneously. It is trained in a simulation combining data-driven modeling techniques with first-principles modeling.
On the one hand, we employ a neural network model to capture the highly nonlinear dynamics of the upper carriage turn hydraulic motor, incorporating explicit pressure prediction to handle delays better. On the other hand, we model the arm as velocity-controllable and the free-swinging end-effector tool as a damped pendulum using first principles. This combined model enhances our simulation environment, enabling the training of RL controllers that can be directly transferred to the real machine. 
Designed to reach steady-state Cartesian targets, the RL controller learns to leverage the hydraulic dynamics to improve accuracy, maintain high speeds, and minimize end-effector tool oscillations. Our controller, tested on a mid-size prototype material handler, is more accurate than an inexperienced operator and causes fewer tool oscillations. It demonstrates competitive performance even compared to an experienced professional driver. 
\end{abstract}


\section{Introduction}

\begin{figure}
    \centering
    \includegraphics[trim={3.5cm 14cm 5cm 4cm},clip, width=\columnwidth]{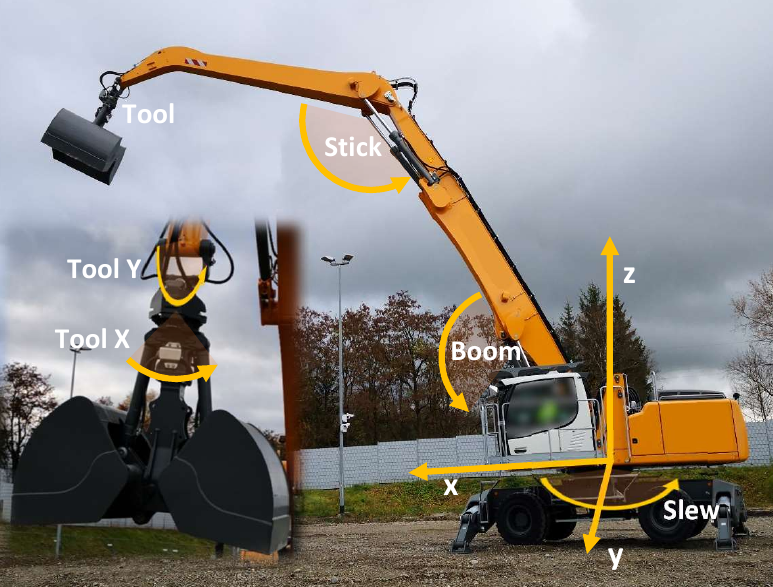}
    \caption{The prototype material handler used in this work has an operational range of about 20~\si{\meter} and weighs more than 40~\si{\tonne}. A 1.5~\si{\tonne} grabbing shovel designed for loose material was employed, with a maximum load of 2~\si{\tonne}.
    }
    \label{fig:machine_frame}
    \vspace{-0.5cm}
\end{figure}

Material handlers similar to the one in \cref{fig:machine_frame} find applications in diverse settings, including construction sites, recycling centers, ports, and warehouses. They are indispensable for efficiently maneuvering and sorting heavy materials such as scrap metal, bulk cargo, logs, and construction debris.
Their most notable feature is the free-swinging end-effector tool. Compared to fixed attachments, it offers reduced manufacturing costs and operational advantages: gravity alignment facilitates grabbing piled material, and the swinging can be exploited to enlarge the reachable task space.
While beneficial for specific tasks, this joint setup, combined with the hydraulic actuation, makes maneuvering extremely complex, even for trained operators. The hydraulic cabin-rotation motor is often characterized by extensive delays and binary braking dynamics, making accurate motion control, particularly stopping, challenging. Furthermore, if the tool oscillations are not adequately damped, they can cause severe damage.

Construction has become a focal point for robotic research in recent years~\cite{chen2018construction}, encompassing areas such as force control~\cite{jud2019autonomous, koivumaki2015stability}, full arm motion control~\cite{egli2022general, lee2022precision}, motion planning~\cite{jelavic2023lstp, lee2021real}, and state estimation~\cite{nubert2022graph}.
Autonomous high-level tasks have been demonstrated, including earth-moving planning for bulldozers~\cite{hirayama2019path} and excavators~\cite{terenzi2023autonomous, zhang2021autonomous}, or rock wall construction~\cite{johns2023framework}. However, past research has rarely addressed the fast and efficient handling of material or the automation of large material handlers as the one we focus on, despite the long-recognized importance of material handling tasks in the industry~\cite{skibniewski1992robotic}. 
Automating these machines would enable continuous operation and limit the need for human drivers in harsh conditions, thereby improving both efficiency and safety.

In this work, we propose a solution for autonomous large-scale material manipulation, filling the gap in the literature of control strategies for material handlers. Most state-of-the-art methods do not work reliably on such machines due to the complex actuators' velocity profile and the lack of specialized high-bandwidth servo valves.
We present a \ac{RL} control scheme for material handling machines, learning to reach 3D Cartesian targets with high speeds while minimizing oscillations to facilitate safe material grasping. The proposed approach combines data-driven modeling for the highly non-linear and delayed cabin turn joint with modeling from first principles for the arm and the free-swinging end-effector tool. 
The evaluated controller performs at a speed and accuracy comparable to those of an average operator and works under any load.

\subsection{Related Work}

\subsubsection{Control of Hydraulic Machines}
A variety of techniques for the control of hydraulic machines have been proposed.
Traditional model-based control approaches~\cite{mattila2017survey, yamamoto2024position} rely the most on an accurate, often even analytic, model to handle delays and non-linearities, which is hard to obtain in practice. Research has shown that retrofitted high-performance hydraulic valves enable accurate force control~\cite{jud2019autonomous}. Their practical use, however, is restricted by their high price and limited oil flow, often only allowing for slow arm motions.
As a response, recent research has focused on incorporating machine learning techniques into the control loop.
Nurmi et al.~\cite{nurmi2018neural} work at the intersection of \ac{ML} and model-based control, proposing a deep learning-based method to identify nonlinear velocity \ac{FF} curves for pressure-compensated hydraulic valves. \ac{FF}-based velocity control is also adopted in our work to control the two arm joints, even though without these learning improvements. 
Park et al.~\cite{park2016online} present an online learning framework for position control of hydraulic excavators using echo-state networks. From time series of input and output data, they train an inverse model of the plant, which is later used to generate the control commands given a new trajectory reference. Online learning is, however, dangerous for heavy machinery tasks, so Lee et al.~\cite{lee2022precision} propose a similar model inversion control approach working offline. In particular, they decompose the model into three physics-inspired components to learn the excavator dynamics more efficiently, considering force and pressure measurements. 
Recently, \ac{RL} has emerged as an alternative to control hydraulic excavation machines. Egli and Hutter~\cite{egli2020towards, egli2022general} introduce a data-driven modeling approach of the coupled hydraulic cylinders. The excavator is controlled by training an \ac{RL} agent in simulation, aiming for end-effector position- or velocity-tracking in free space and with weak ground contact. Compared to previous classical approaches, the learned controller handles non-linear dynamics and delays better and is more robust to disturbances. 
In this work, we build on top of~\cite{egli2022general} but split the modeling into two parts: \textit{i)} the data-driven analogy is used for the cabin turn joint, with pressure and inertia as additional features, \textit{ii)} while first principles modeling is adopted for the others. 
Dhakate et al.~\cite{dhakate2022autonomous} propose a similar pipeline for a different machine. Their work captures the mapping between cylinders' displacements and joint variables of a small forest forwarder crane through a \ac{NN} model. They then train an \ac{RL} position controller, which commands joint setpoints, simply treating the unactuated tool joint as a disturbance. In contrast, we explicitly model the unactuated tool in our simulation and aim for an active damping behavior through suitably chosen actions of all the controllable joints.

\subsubsection{Control in the Presence of Passive Joints}
Previous research in the construction domain has rarely addressed the free-swinging end-effector tool. 
Promising results on the safe control of tower cranes~\cite{ramli2017control, rauscher2020modeling} have been achieved, but they a have simpler structure and more restricted tasks. \ac{RL} has been used on these machines to improve control performance under payload variation~\cite{zhang2023deep}.   
Andersson et al.~\cite{andersson2021reinforcement} use \ac{RL} to control a simulated forestry crane for log grasping. The arm and grapple kinematics resemble those of material handlers, and the agent learns to take advantage of the oscillations to complete the task. However, the motors are assumed to track velocity references reliably on every joint, and the work is not validated on physical hardware.
Oktay and Sultan~\cite{oktay2013modeling} explore the helicopter slung-load system, modeled using first principles. Simulation results show the dependence of the model-based controller on the exact dynamic parameters to operate reliably. Further studies have focused on trajectory optimization~\cite{sreenath2013quadrotor} and \ac{RL}~\cite{palunko2013suspended}, addressing tracking with the suspended load by enabling the aerial system to exploit inertial forces for motion generation. 
In the robotic manipulation domain, Zimmermann et al.~\cite{zimmermann2019puppetmaster} developed a computational framework for the robotic animation of string puppets. These are coupled pendulum systems, sharing similar dynamics with a free-swinging grab. 
Ichnowski et al.~\cite{ichnowski2022gomp} work on inertial transport for pick-and-place operations with robotic arms. Their approach is based on iterative convex optimization with end-effector acceleration constraints. Fictitious forces are included in the model to consider the inertial load during planning. Our work applies similar modeling techniques to build the training environment but solves the control problem via \ac{RL}.

\vspace{-0.1cm}
\subsection{Contributions}
We present the following contributions:
\begin{itemize}
    \item A data collection routine and an \ac{NN}-architecture to model hydraulic motors, accurately predicting delay effects by leveraging velocity and pressure evolutions.
    \item A combined modeling approach, partly consisting of a data-driven \ac{NN}-model for the slew, partly of first-principle modeling for the arm and the unactuated tool.
    \item An \ac{RL} agent entirely trained in simulation using the previously mentioned mixed model, actuating the slew motor and the two arm joints simultaneously. 
    The controller tracks steady-state Cartesian position targets on the real machine, effectively reducing the tool oscillations while maintaining a high operational speed. 
    \item Experimental validation on a prototype 40~\si{\tonne} material handler, including comparisons with human operators of varying experience levels.
\end{itemize}
\section{System Description}

Our algorithm was validated on real hardware~(\cref{fig:machine_frame}). This research machine has three hydraulically actuated joints with one~\ac{DoF} each. They are independently controlled via steer-by-wire joystick commands and are subject to delays, dead zones, and non-linearities. Linear hydraulic cylinders actuate the boom and stick joints, while the slew joint uses a hydraulic motor to rotate freely. This motor presents a binary braking system to slow down rotations more aggressively, but its contribution cannot be actively regulated. The relationship between pressure and velocity is highly intertwined, and the configuration's inertia plays a crucial role in shaping the speed curve. Such actuation complexity renders the previously deployed model-based control algorithms unreliable, resulting in significant overshoots and oscillations. Consequently, we use an \ac{ML} approach for both the slew modeling and control. 
The tool consists of a chain of two unactuated revolute joints (pitch \& roll) and one actuated joint for rotations around the vertical axis (yaw). Since our controller solely manages gripper positioning, we disregard the last joint as well as the clamshell opening, reducing the overall kinematics of the system to five \acp{DoF}.

\vspace{-0.1cm}
\subsection{Feedback}
The machine used, specifically developed for autonomous purposes, is retrofitted at 50Hz with:
\begin{itemize}
    \item Encoders on slew, boom, and stick joints, providing position and velocity measurements. As the velocity is derived from position data, it suffers from a delay of approximately 0.2~\si{\second}.
    \item Pressure sensors on slew, boom, and stick, measuring fluid pressure on both sides of the piston.
    \item \acp{IMU} on cabin and tool, each supplying 3D angular velocities, 3D linear accelerations, and 2D angular orientation. 
    \item An algorithm estimating the material load in the tool using least-squares regression on pressure data. 
\end{itemize}

\vspace{-0.1cm}
\subsection{Arm Velocity Controller}
A model-based velocity controller for the arm joints has been formulated, incorporating a \ac{FF} component and PI feedback compensation as discussed in~\cite{jud2021heap}. The \ac{FF} model is based on a 25-point \ac{LUT} per joint and maps desired joint velocities to control commands. 
Such an approach can be deployed for simple isolated hydraulic cylinder control. 
While this kind of direct control cannot be applied on the slew hydraulic motor, it is tuned well enough for boom and stick simple hydraulic cylinders, allowing for a convenient decoupling of arm motion planning and low-level cylinder control as in~\cite{egli2022soil}.

\vspace{-0.1cm}
\section{Proposed Approach}
Our research centers around the following aspects: \textit{i)} utilizing \ac{ML} to model hydraulic motor dynamics, \textit{ii)} constructing a simulation environment with a reduced sim-to-real gap, and \textit{iii)} training an \ac{RL} controller to track task-space targets with the arm while accommodating hydraulic dynamics and tool oscillations.

\vspace{-0.1cm}
\subsection{Slew Actuator Model}
We use \ac{ML} to capture the turning dynamics as a first step toward RL-based slew control.

\subsubsection{Data Collection}
We aimed to excite the slew motor's main modes and explore the relevant state space. Data was collected in various ways according to \cref{tab:dataset}. The majority has been generated by applying artificial excitation signals consisting of regular periodic references. For each run, the arm configuration was randomized and kept static. We further collected data during manual random movements and real driving situations. For the final model, we included the deployment of an earlier version of the slew controller, trained only on the first two data sources. By recording a driver during operational cycles and our controller running in a closed loop, the model became more accurate at predicting the state evolution of typical tasks. 
We then built a unique dataset, as fine-tuning and transfer learning~\cite{torrey2010transfer} do not benefit our simple architecture.

\renewcommand{\arraystretch}{1.2}
\begin{table}
    \begin{center}
        \caption{Characteristics of the Training Set.}
        \label{tab:dataset}
        \begin{tabular}{@{}llp{6cm}@{}}
            \toprule
            \textbf{\%} & \textbf{Mins} & \textbf{Description}\\
            \midrule
            70 & 56 & Periodic references: step, sinusoidal, trapezoidal, with static arm. Random references with arm motion. \\
            9 & 7 & Real driving conditions during common operations. \\
            21 & 16 & Closed-loop slew controller, with static arm. \\
            \midrule
            100 & 79 & Total data collected. \\
            \bottomrule 
        \end{tabular}
    \end{center}
    \vspace{-0.5cm}
\end{table}

\subsubsection{Data Augmentation}
From experiments and mechanical analysis, we concluded that the slew joint has symmetric rotation dynamics. Leveraging this, we augmented the dataset via mirroring, which doubles the amount of data and improves the model accuracy due to the attenuation of the recorded noise effects.

\subsubsection{Neural Network Model}\label{sec:NN_model}

\renewcommand{\arraystretch}{1.2}
\begin{table}[b]
    \vspace{-0.5cm}
    \begin{center}
         \caption{Neural Network Design Hyperparameters.}
         \label{tab:nn_params}
         \begin{tabular}{@{}lcc@{}}
             \toprule
             \textbf{Hyperparam} & \textbf{Pressure} & \textbf{Velocity} \\ \midrule
             Input dim & 41 & 41 \\
             Output dim & 2 & 1 \\
             Layers & $\big[ 128, 128, 128, 128, 32 \big]$ & $\big[ 128, 128, 128, 32 \big]$ \\
             \bottomrule
         \end{tabular}
    \end{center}
    \vspace{-0.5cm}
\end{table}

We use a \ac{MLP} to capture the slew dynamics. The framework operates on a vectorized input history of states and control commands and predicts the pressure and velocity evolution one step into the future.
Unlike previous work~\cite{egli2022general}, we explicitly include measured pressures in our formulation to better deal with the large delays from command input to velocity response. In particular, we learn both pressure (\cref{eq:NN_formulation_press}) and speed (\cref{eq:NN_formulation_vel}) dynamics via two different \acp{MLP}, with trainable weights $\theta_p$ and $\theta_\omega$. The prediction and measurement rates are $0.1$~\si{\second}. Both use the ReLU activation function inspired by~\cite{nurmi2018neural}, with the loss defined as the single-step prediction error. Other hyperparameters are summarized in \cref{tab:nn_params}. 
We assume a deterministic mapping between the past inputs and the next pressure values and model it as:
\begin{equation}
\label{eq:NN_formulation_press}
    \begin{aligned}
        \Big[p_{l[k]}, p_{r[k]} \Big] = \mathcal{F}_p\Big(
        &u_{[k-9,k]}, p_{l[k-10,k-1]}, p_{r[k-10,k-1]}, \\ 
        & \omega_{[k-10,k-1]}, I_{z,[k-1]}; \theta_p \Big),
    \end{aligned}
\end{equation}
where $\mathcal{F}_p$ denotes the neural network with trainable parameters $\theta_p$, $u$ denotes the control input, $p_l$ and $p_r$ represent the pressures of the left and right chambers, $\omega$ is the angular velocity of the cabin and $I_z$ the configuration-dependent inertia around the z-axis. The notation $\cdot[i,j]$ denotes a discrete time series from time step $i$ to time step $j$ of the given quantity. Using our double architecture, 1~\si{\second} history access is enough for generalizable learning.   
Similarly, the second neural network models the velocity evolution as follows:
\begin{equation}
\label{eq:NN_formulation_vel}
    \begin{aligned}
        \Big[\omega_{[k]} \Big] = \mathcal{F}_\omega \Big( 
        u_{[k-9,k]},& p_{l[k-9,k]}, p_{r[k-9,k]}, \\ 
        & \omega_{[k-10,k-1]}, I_{z,[k-1]}; \theta_\omega \Big) .
    \end{aligned}
\end{equation}
Note that \cref{eq:NN_formulation_vel} takes the output of \cref{eq:NN_formulation_press} as an input. Finally, the position at step $k$ is computed via integration. All buffers are initialized with zeros.
The configuration-dependent inertia $I_z$ is computed from the arm position and the nominal link weights, approximating the tool as a point mass. This simple approach is sufficient to capture the effects of the arm extension on the acceleration. We report in \cref{tab:nn_compare} the \ac{MAE} obtained with different input features but the same history access. Notably, inertia plays the most critical role, and the additional pressure inputs are useful to better learn the velocity dynamics. 

\begin{figure}
    \centering
    \includegraphics[trim={0cm 0cm 0cm 0cm},clip,width=\columnwidth]{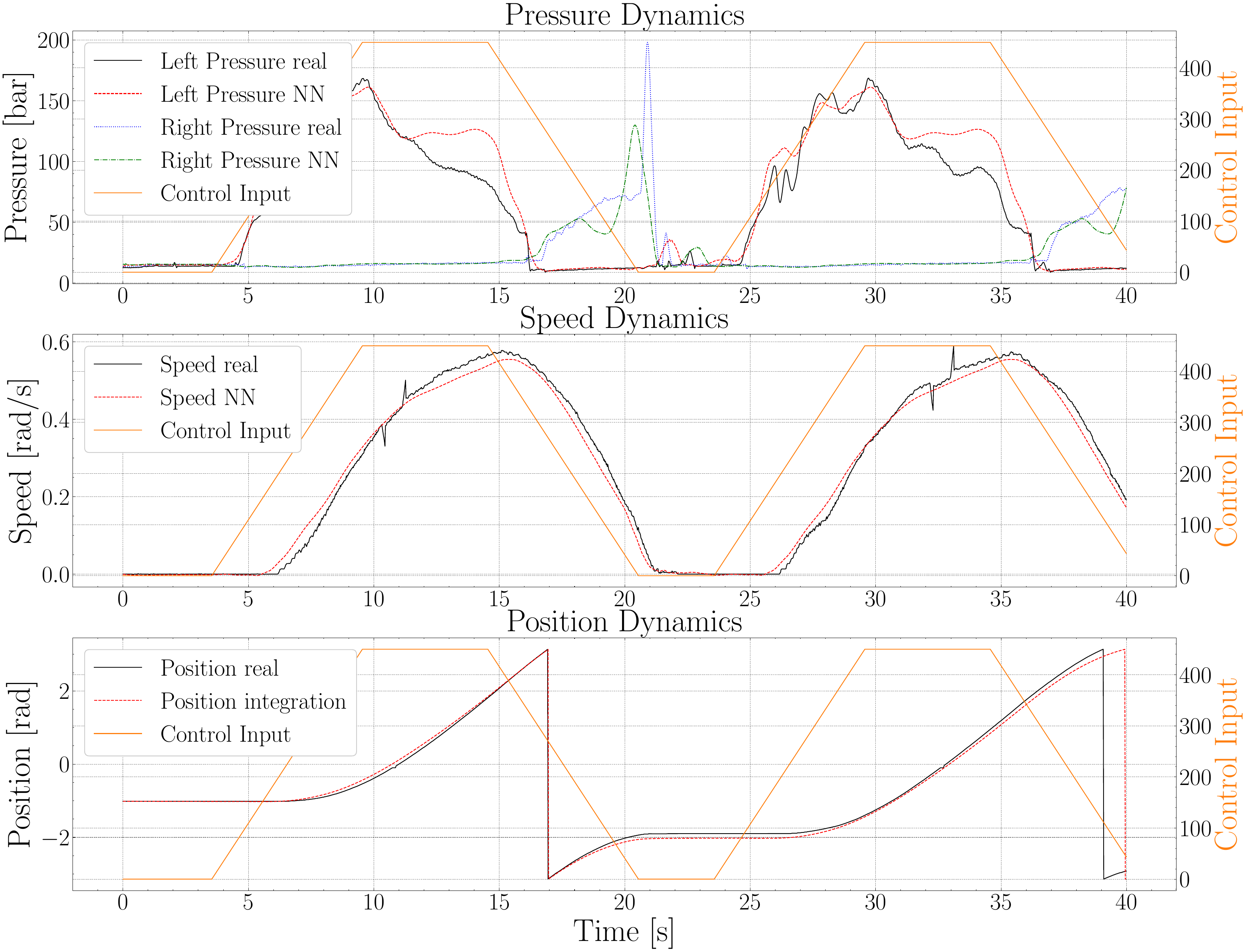}
    \caption{Open-loop prediction using the \ac{NN} model for a 40-second trapezoidal reference. This shape approximates a control profile while maintaining regularity to mitigate noise effects.}
    \label{fig:open_loop_pred}
    \vspace{-0.3cm}
\end{figure}

\cref{fig:open_loop_pred} shows the qualitative open-loop performance of our model on a $40$~\si{\second}-long test trajectory. The integrated position is subject to drift over time. Still, we must ensure accuracy only for a training episode, during which the \ac{RL} agent achieves its goal. Based on real-world manipulation tasks, we identify $10$~\si{\second} as a good trade-off between sufficiently long and precise.

\renewcommand{\arraystretch}{1.2}
\begin{table}[b]
    \begin{center}
         \caption{Actuator Model Feature Crafting.\\ Metrics are computed over $10$\si{\second} open-loop predictions.}
         \label{tab:nn_compare}
         \begin{tabular}{@{}lcc@{}}
             \toprule
             \textbf{Features} & \textbf{Velocity MAE} [\si{\degree}/\si{\second}] & \textbf{Position MAE} [\si{\degree}] \\ \midrule
             Proposed & \textbf{1.616} &  \textbf{5.707} \\
             No Pressure & 1.805 & 6.509 \\
             No Inertia & 2.080 & 7.517 \\
             \bottomrule
         \end{tabular}
    \end{center}
    \vspace{-0.3cm}
\end{table}

\vspace{-0.1cm}
\subsection{Simulation Environment}

\begin{figure}
    \centering
    \begin{minipage}{0.51\columnwidth}
       \includegraphics[trim={0cm 0cm 0cm 0cm},clip,width=\textwidth]{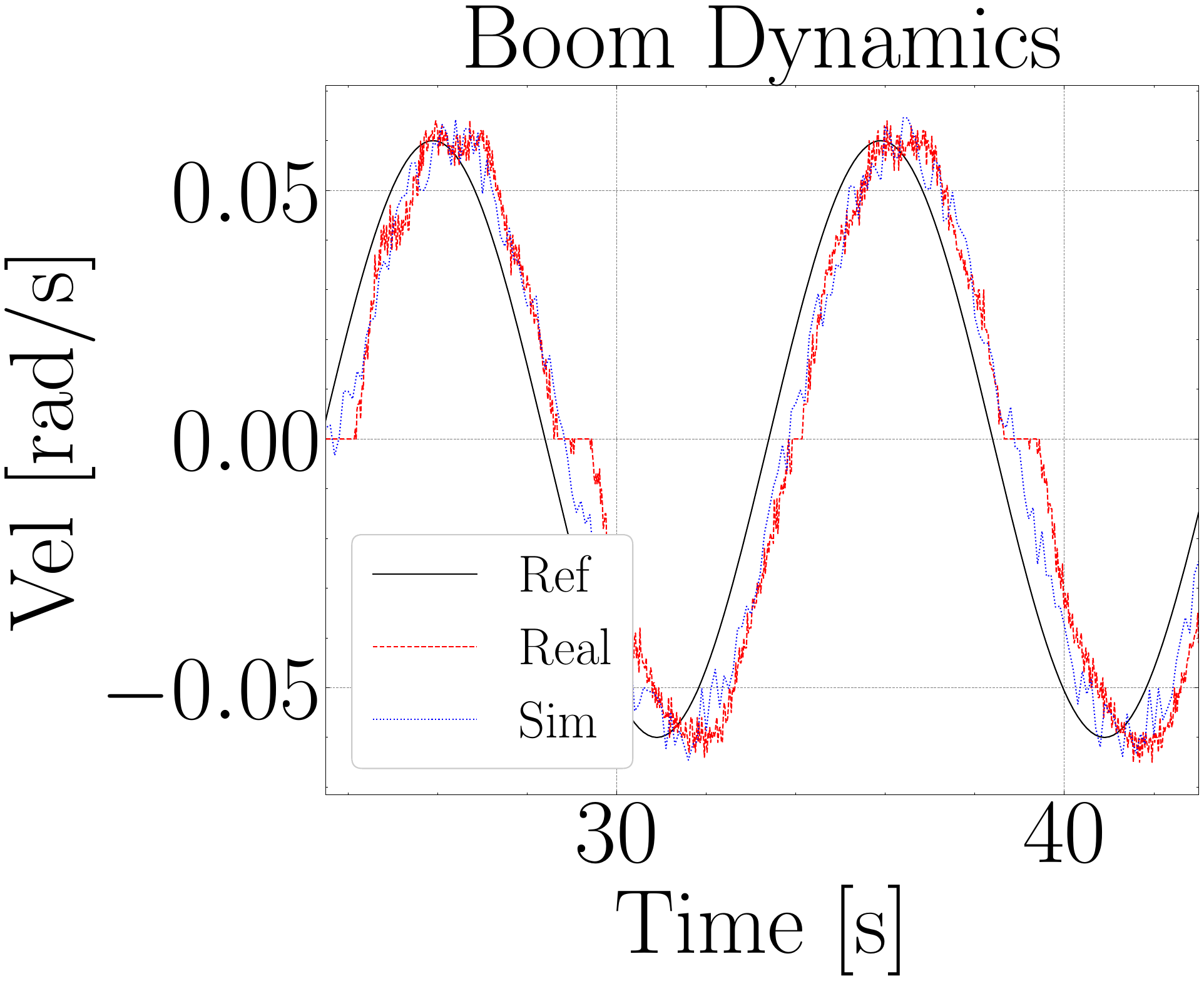} 
    \end{minipage}
    \begin{minipage}{0.47\columnwidth}
       \includegraphics[trim={0cm 0cm 0cm 0cm},clip,width=\textwidth]{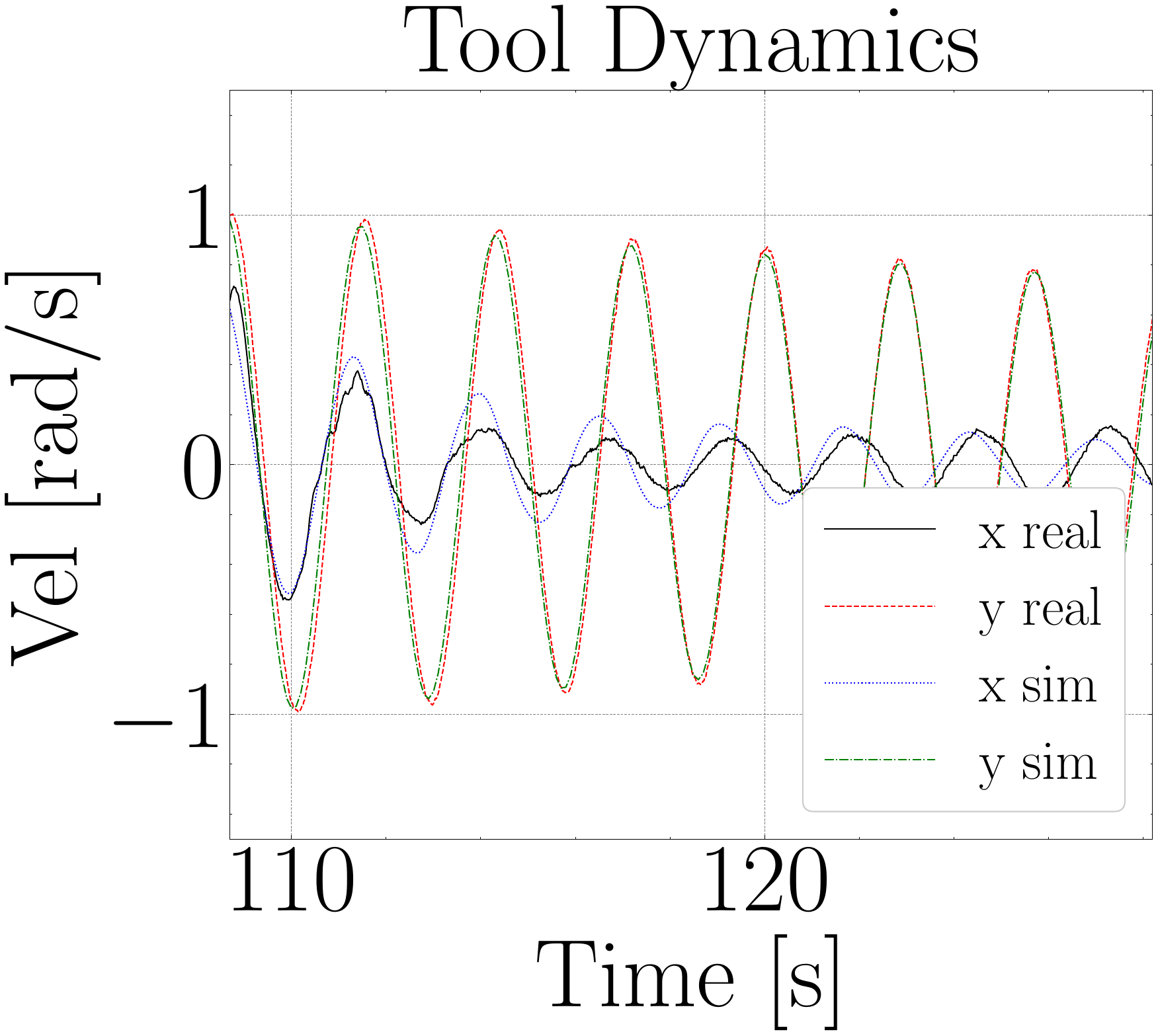} 
    \end{minipage}
    \caption{Arm controller dynamics modeled as first-order systems with delay (left), and tool dynamics modeled via Lagrange and dissipation (right).}
    \label{fig:ee_control_env}
    \vspace{-0.5cm}
\end{figure}

A mixed environment was developed to train \ac{RL} controllers.
It simulates the full machine \textit{i)} using the actuator model for the slew joint dynamics and \textit{ii)} using an analytic model derived from first principles to simulate the arm and the tool dynamics. While the first is machine-specific, the mathematical models can be used more generally and adapted with parameter tuning. 
For the boom and stick joints, we use a first-order system with delay to approximate the velocity tracking performance of the arm controller:
\begin{equation}\label{eq:arm_simulated}
    \Dot{q}[k] = \Dot{q}[k-1] + P\big( \Hat{\Dot{q}}[k - d] - \Dot{q}[k-1] \big) ,
\end{equation}
where $\Hat{\Dot{q}}$ is the velocity reference, $d$ represents the delay (larger than the one produced by the encoders), and $P$ specifies a hand-tuned time constant. This equation produces simulated trajectories as in \cref{fig:ee_control_env}.

\begin{figure}
    \centering
    \begin{minipage}{0.5\columnwidth}
    \includegraphics[trim={5.5cm 6cm 8cm 10.5cm},clip, width=\textwidth]{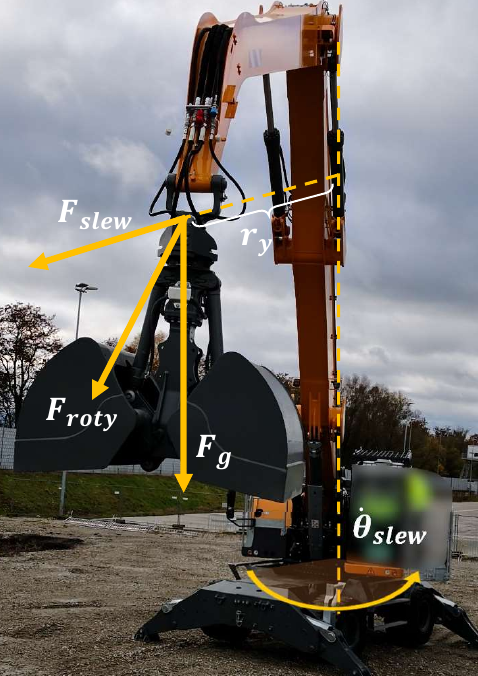}
    \end{minipage}\hfill
    \begin{minipage}{0.5\columnwidth}
    \includegraphics[trim={8cm 8cm 7.5cm 11.2cm},clip,width=.9\textwidth]{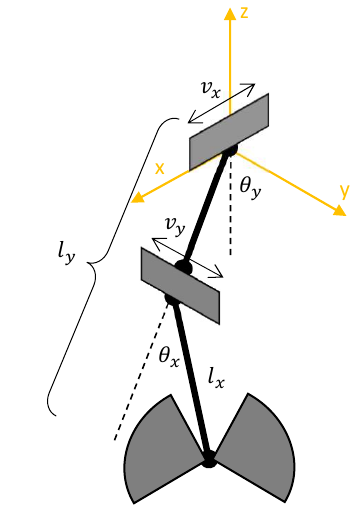}
    \end{minipage}
    \caption{The tool is modeled as a double pendulum with linearly oscillating support. In the left figure, we show the forces accounted for. The adopted approximations for each \ac{DoF} are shown on the right.
    }
    \label{fig:tool_all}
    \vspace{-0.5cm}
\end{figure}

The tool is modeled using the Lagrange principle as a pendulum with decoupled $x$ and $y$ rotations. This makes state propagation via Forward Euler integration ($\Delta t = 0.02$~\si{\second}) easier and more stable. The moving reference frame produces fictitious forces; we account only for the Euler and the centrifugal ones, as the Coriolis force would introduce a coupling between the axes. 
Furthermore, we include a dissipative term with the Rayleigh's dissipation function~\cite{minguzzi2015rayleigh}. These choices lead to a system response as illustrated in \cref{fig:ee_control_env}.
The following equations describe the mathematical formulation:
\begin{equation}\label{eq:tool_y}
    \begin{aligned}
        \dot{\theta}_{y[k+1]} &= \Bigg(\bigg(\frac{v_{x[k+1]}-v_{x[k]}}{\Delta t}\cos \theta_{y[k]} -\underbrace{g\sin \theta_{y[k]}}_{F_g} \\ &- \underbrace{\dot{\theta}_{slew[k]}^2r_y}_{F_{slew}}\bigg)/l_y - \underbrace{b_{fy}\dot{\theta}_{y[k]}}_\text{dissipation}\Bigg)\Delta t + \dot{\theta}_{y[k]},
    \end{aligned}
\end{equation}
\begin{equation}\label{eq:tool_x}
    \begin{aligned}
        \dot{\theta}_{x[k+1]} &= \Bigg(\bigg(-\frac{v_{y[k+1]}-v_{y[k]}}{\Delta t}\cos \theta_{x[k]} -\big(\underbrace{g\cos \theta_{y[k]}}_{F_g} \\ &+ \underbrace{\dot{\theta}_{y[k]}^2l_y}_{F_{roty}}\big)\sin \theta_{x[k]}\bigg)/l_x - \underbrace{b_{fx}\dot{\theta}_{x[k]}}_\text{dissipation}\Bigg)\Delta t + \dot{\theta}_{x[k]} .
    \end{aligned}
\end{equation}
Here, $\theta_{x,y}$ denote the angles between the two unactuated joints and the main frame axes, $v_{x,y}$ the linear velocities of the tool attachment points, $l_{x,y}$ the corresponding tool lengths, $r_y$ the distance between the tool and the slew rotation axis, $g$ the gravity constant, and $b_{fx,y}$ the dissipative coefficients. An illustration is shown in \cref{fig:tool_all}.

\vspace{-0.1cm}
\subsection{RL End-Effector Controller}
Inspired by the prior success of learning-based control of hydraulic excavators~\cite{egli2020towards,egli2022general}, we tackled the end-effector task-space control by training a high-level \ac{RL} policy that \textit{i)} deals with the nonlinear machine dynamics (mostly stemming from the hydraulic slew motor), and \textit{ii)} actively stabilizes the tool oscillations. 
This trained policy directly outputs joystick commands for the slew joint and speed references for boom and stick low-level velocity controllers. 

\subsubsection{Control Formulation}
The controller is represented by an \ac{MLP} with trainable weights $\theta_c$ as follows:
\begin{equation}\label{eq:full_policy}
    \begin{aligned}
        \mathbf{u}_{[k]} & = \Big[u_{\text{slew}[k]},\Hat{\Dot{q}}_{\text{boom}[k]},\Hat{\Dot{q}}_{\text{stick}[k]}
        \Big]^\top \\
        & = \pi_{\theta_c} \Big( \mathbf{u}_{[:]}, \Dot{\mathbf{q}}_{[:]}, \mathbf{q}_{[:]}, I_{z,[k-1]}, \Hat{x},\Hat{y},\Hat{z} \Big), \\
    \end{aligned}
\end{equation}
where the buffer is $[:] \coloneqq [k-H, k-1]$, and
\begin{equation*}
    \begin{aligned}
        &\mathbf{q} = \big[q_{\text{slew}}, q_{\text{boom}}, q_{\text{stick}}, q_{\text{tool},x}, q_{\text{tool},y} \big]^\top, \\
        &\Dot{\mathbf{q}} = \big[\Dot{q}_{\text{slew}}, \Dot{q}_{\text{boom}}, \Dot{q}_{\text{stick}}, \Dot{q}_{\text{tool},x}, \Dot{q}_{\text{tool},y} \big]^\top, \\
    \end{aligned}
\end{equation*}
denote the measured angular position and velocity of all joints.
Given a history of actions $\mathbf{u}$ and states $[\mathbf{q},\Dot{\mathbf{q}}]$, the inertia $I_z$, and the task-space target $[\Hat{x},\Hat{y},\Hat{z}]$, the \ac{RL} controller produces $u_{slew}$ (directly applied to the machine), $\Hat{\Dot{q}}_{boom}$ and $\Hat{\Dot{q}}_{stick}$ (arm velocity references). 
During training, the Cartesian target is constant throughout an entire $10$~\si{\second} episode.
The joystick input is limited to feasible values, and speed references are clipped within $[-0.2,0.2]$~\si{\radian}/\si{\second} to match the \ac{LUT} steady-state velocity assumption.
The history length $H$ is a trade-off between learning to exploit the full range of dynamics and limiting the mismatch of simulation and reality, which increases with $H$. 

\subsubsection{Policy Gradient RL}
We train the agent using model-free policy gradient \ac{RL}. 
Specifically, we employ the PPO~\cite{schulman2017proximal} learning scheme with 50~\si{\hertz} simulation and 10~\si{\hertz} control rate. Our policy and value function networks have dimensions of $[256, 128, 128]$, using $\tanh$ activation and a linear output layer.

\subsubsection{Domain Randomization}
We use domain randomization to bridge the sim-to-real gap~\cite{zhao2020sim, hwangbo2019learning}. 
At each step, uniform noise is added to the observations. At each environment initialization, we randomly sample the initial joint positions (avoiding collision configurations), the load, the arm controller and tool model parameters, and the target. Parameter randomization is particularly important for control robustness during deployment, and helps alleviate the simulation inaccuracies.    
The tool starts vertically, with zero initial velocity. Based on the assumption that the slew joint behavior is independent of position, the Cartesian target is randomly generated only in the $x$-$z$ plane, with $\hat{y}=0$ and $\hat{\theta}_{\text{slew}}=0$. This facilitates learning by reducing the number of active observations. When deploying for different targets, the slew position feedback is then converted to an error.

\subsubsection{Termination Conditions}
Besides the low-level velocity controller enforcing boom and stick position limits, we also include termination conditions during training to avoid any safety hazards, i.e., if the gripper gets too low or too close to the cabin. 
This way, self-collisions and collisions with flat ground are avoided precautionary. 
All episodes last $10$~\si{\second}, unless the mentioned termination occurs. 

\subsubsection{Reward}
Our proposed reward at timestep $k$ consists of seven terms: 
\begin{equation}\label{eq:reward_full}
    \begin{aligned}
        R_k & = r_k^{\text{balance}} + r_k^{\text{target}} + r_k^{\text{action}} \\
        & + r_k^{\text{overshoot}} + r_k^{\text{oscillation}} + r_k^{\text{decouple}} + r_k^{\text{one-shot}}.
    \end{aligned}
\end{equation}
These are defined as follows:
\begin{equation*}
    \begin{aligned}
        r_k^{\text{bal.}} & \propto \Big(\exp\big(-\| \Tilde{\varepsilon}_k \|_1\big) -1\Big), \quad 
        r_k^{\text{tar.}} \propto \exp\Big(-\| \Tilde{\varepsilon}_k \|_2^2\Big), \\
        r_k^{\text{act.}} & \propto -\| \Delta \textbf{u}_k / \sigma_u \|_2^2, \quad
        r_k^{\text{over.}} \propto \Big(\exp\big(-|q_k^{\text{ovs}}|\big)-1\Big), \\
        r_k^{\text{osc.}} & \propto -\| \Tilde{\varphi}_k \|_1, \quad 
        r_k^{\text{dec.}} \propto -|\Dot{q}_{\text{slew}[k]}| \big( |\Dot{q}_{\text{boom}[k]}| + |\Dot{q}_{\text{stick}[k]}| \big), \\
        r_k^{\text{o-s}} & \propto -\| \textbf{u}_k / \sigma_u \|_2^2 \cdot \mathcal{I} \big[ \| \Tilde{\varepsilon}_k \|_2 < 0.5 \wedge |\Dot{q}_{\text{slew}[k]}| < 0.02 \big]
    \end{aligned}
\end{equation*}
with quantities
\begin{equation*}
    \begin{aligned}
        \Tilde{\varepsilon}_k & = \big[\Hat{x}-x_{[k]}, \Hat{y}-y_{[k]}, \Hat{z}-z_{[k]} \big], \\
        \textbf{u}_k & = \big[u_{\text{slew}[k]}, \Hat{\Dot{q}}_{\text{boom}[k]}, \Hat{\Dot{q}}_{\text{stick}[k]} \big], \; \Tilde{\varphi}_k = \big[ \Dot{q}_{\text{tool},x[k]}, \Dot{q}_{\text{tool},y[k]} \big] \\
        q_k^{\text{ovs}} & = 
        \begin{cases}
            \max\big( 0, q_{\text{slew}[k]} \big) & \text{if} \quad q_{\text{slew}[0]} < 0, \\
            \min\big( 0, q_{\text{slew}[k]} \big) & \text{if} \quad q_{\text{slew}[0]} > 0
        \end{cases}.
    \end{aligned}
\end{equation*}

\paragraph{Core Reward Terms}
We use the $r^{\text{balance}}$ penalty to promote a fast target approach while limiting its maximum magnitude early in the episode. We also define a positive reward for reaching the target proximity: $r^{\text{target}}$.
Both these terms are scaled and reshaped with curriculum learning~\cite{bengio2009curriculum}, a technique which has become extremely popular in \ac{RL}~\cite{soviany2022curriculum} because it allows agents to acquire complex skills by being tasked with environments of increasing difficulty. Two separate rewards allow for easier and more effective tuning. 
To reduce the aggressiveness of the policy, we introduce $r^{\text{action}}$, computed on the normalized delta action vector. 
\paragraph{Decorating Reward Terms}
Additional reward terms further shape the behavior towards material handling tasks. 
The overshoot penalty $r^{\text{overshoot}}$ allows for additional arm motion to cope with the tool oscillations but enforces that the slew angle target is reached with proper braking.
We penalize the tool velocity oscillations via $r^{\text{oscillation}}$. This term prevents the tool from reaching dangerous configurations during the motion and promotes damping upon target achievement. 
Further, we include a loss, $r^{\text{decouple}}$, which aims to reduce the coupling between slew and arm motions. As the simulated slew actuator model does not consider inertia variations, this term limits model mismatches.
The one-shot penalty $r^{\text{one-shot}}$ prevents additional actions once a target neighborhood has been reached, so that the machine can initiate the grasping phase. 
We slowly introduce it via curriculum to prevent the agent from diverging during the initial training iterations.  

\vspace{-0.2cm}
\subsection{RL Slew-Only Controller}
To validate the \ac{RL}-based approach and assess the quality of the slew joint model, we first conducted tests using policies trained to actuate only the rotational slew motor (without the arm joints). This controller primary objective is to learn how to accurately command the nonlinear rotation dynamics to reach target joint positions, neglecting the tool. We utilized some of the reward components outlined in \cref{eq:reward_full}, specifically $r^{\text{balance}}, r^{\text{target}}, r^{\text{action}}, r^{\text{overshoot}}, r^{\text{one-shot}}$, but adapted to operate in the joint space.

\vspace{-0.1cm}
\section{Evaluation}

\begin{figure}
    \centering
    \includegraphics[trim={2cm 19cm 2cm 3cm},clip,width=\columnwidth]{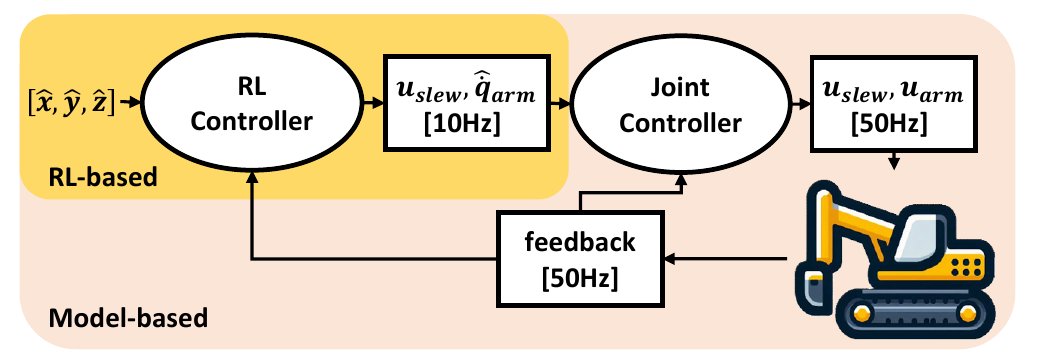}
    \caption{Schematic of the ROS 2 interface. Nodes are oval, and the communication interfaces are represented in rectangular boxes with message rates. 
    The \textit{RL Controller} outputs three actions $\big[u_{slew}, \Hat{\Dot{q}}_{boom}, \Hat{\Dot{q}}_{stick} \big]$ at 10Hz, interpreted by the \textit{Joint Controller} to provide arm joystick inputs at 50Hz, using \ac{FF} and PI compensation, while maintaining a constant zero-order hold slew joystick signal for 5 iterations.}
    \label{fig:ros2_ctrl}
    \vspace{-0.3cm}
\end{figure}

We deployed the controllers on a prototype machine using a ROS 2 interface as shown in \cref{fig:ros2_ctrl}.
Tests consist of steady-state position references with transitions every $15$~\si{\second}, similar to typical working routines. We compared policies with differently-tuned rewards and observation vectors to human operators under variable load conditions.
The method is benchmarked using four metrics: \textit{i)} average slew speed until steady-state, \textit{ii)} average of the maximum slew overshoots and \textit{iii)} of the errors at steady state for each target position, and \textit{iv)} average of the tool's angular velocity. These are chosen to validate the suitability of our controller for a material dumping routine, which needs to be efficient but also satisfy hard error constraints. Specifically, the grab needs to approach a ship or a dump truck reliably without collisions. 

\vspace{-0.2cm}
\subsection{Slew-Only Control}

Our slew controller incorporates a five-element history access, allowing the agent to learn how to shape the system dynamics with smooth actions and minimal corrections.
As shown in \cref{tab:slew_control}, history makes the controller faster and more accurate, with an average steady-state error of less than 2\si{\degree}. 
The \ac{RL} approach can accurately control the slew joint, whereas a model-free PI controller fails. However, this simplified formulation does not address the goal of minimizing the tool oscillations. 
Although a slower and smoother rotation is helpful, active damping with all joints is necessary to achieve competitive operation speed.

\begin{table}[t]
    \begin{center}
         \caption{RL Performance for Slew-only Control. \\ As a baseline, we use a PI controller fine-tuned to quickly reduce the error within the $15$~\si{\second} target period. We test 2 \ac{RL} controllers with and without observations history.}
         \label{tab:slew_control}
         \begin{tabular}{@{}ccccc@{}}
             \toprule
             \textbf{Policy} & \textbf{Speed} [\si{\degree}/\si{\second}] & \textbf{Overshoot} [\si{\degree}] & \textbf{Error} [\si{\degree}] & \textbf{Tool} [\si{\degree}/\si{\second}] \\ \midrule
             PI control & 14.04 & 38.10 & 12.38 & - \\
             No History & 11.75 & 6.42 & \textbf{0.46} & 36.50 \\
             History & \textbf{14.44} & \textbf{2.98} & 1.43 & \textbf{25.44} \\ \bottomrule
         \end{tabular}
    \end{center}
    \vspace{-0.5cm}
\end{table}

\vspace{-0.1cm}
\subsection{Full End-Effector Control}

\begin{figure}
    \centering
    \includegraphics[trim={2cm 2.5cm 0.5cm 0.5cm},clip, width=.9\columnwidth]{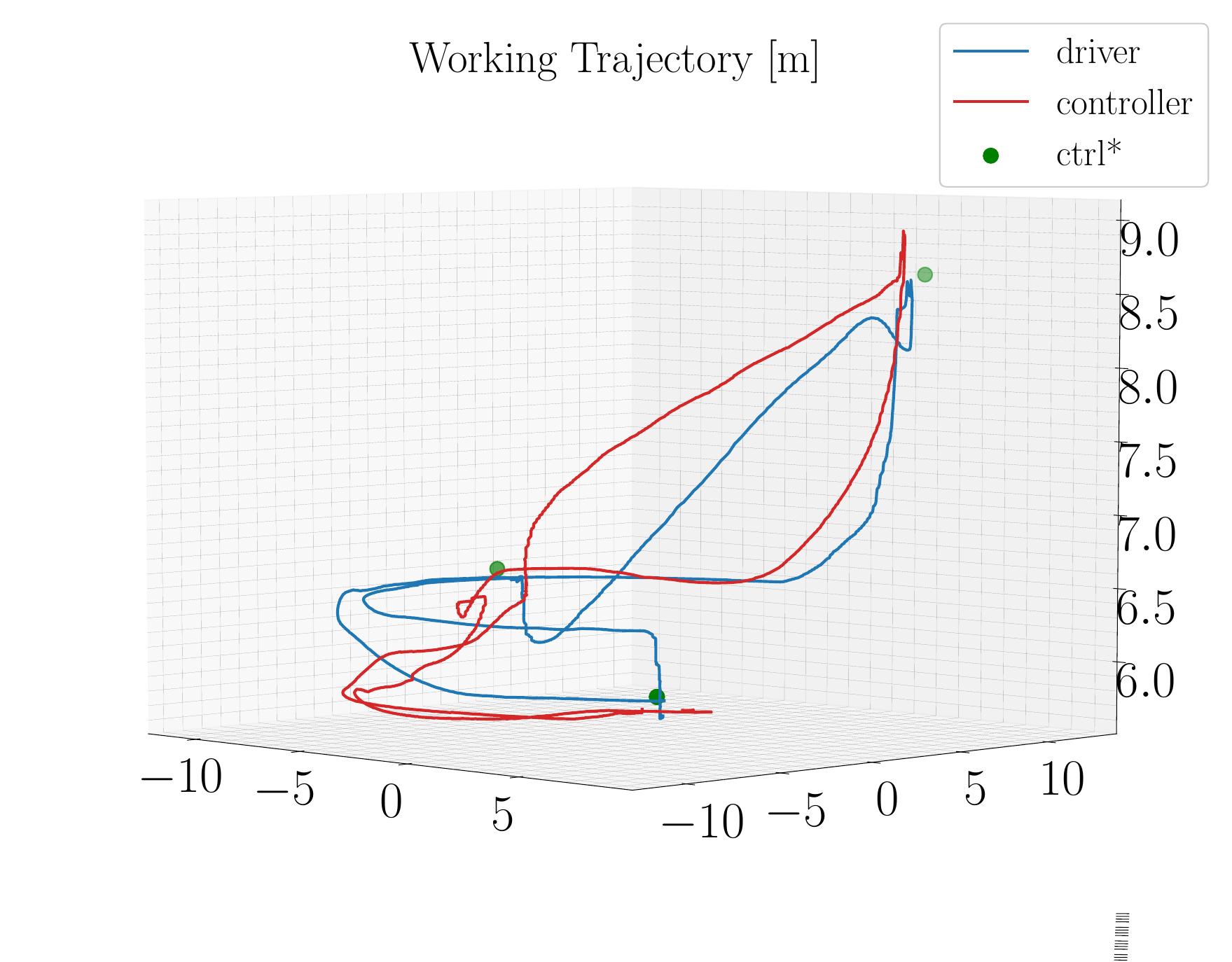}
    \begin{minipage}{0.48\columnwidth}
    \includegraphics[trim={0cm 0cm 0cm 0cm},clip, width=.8\textwidth, right]{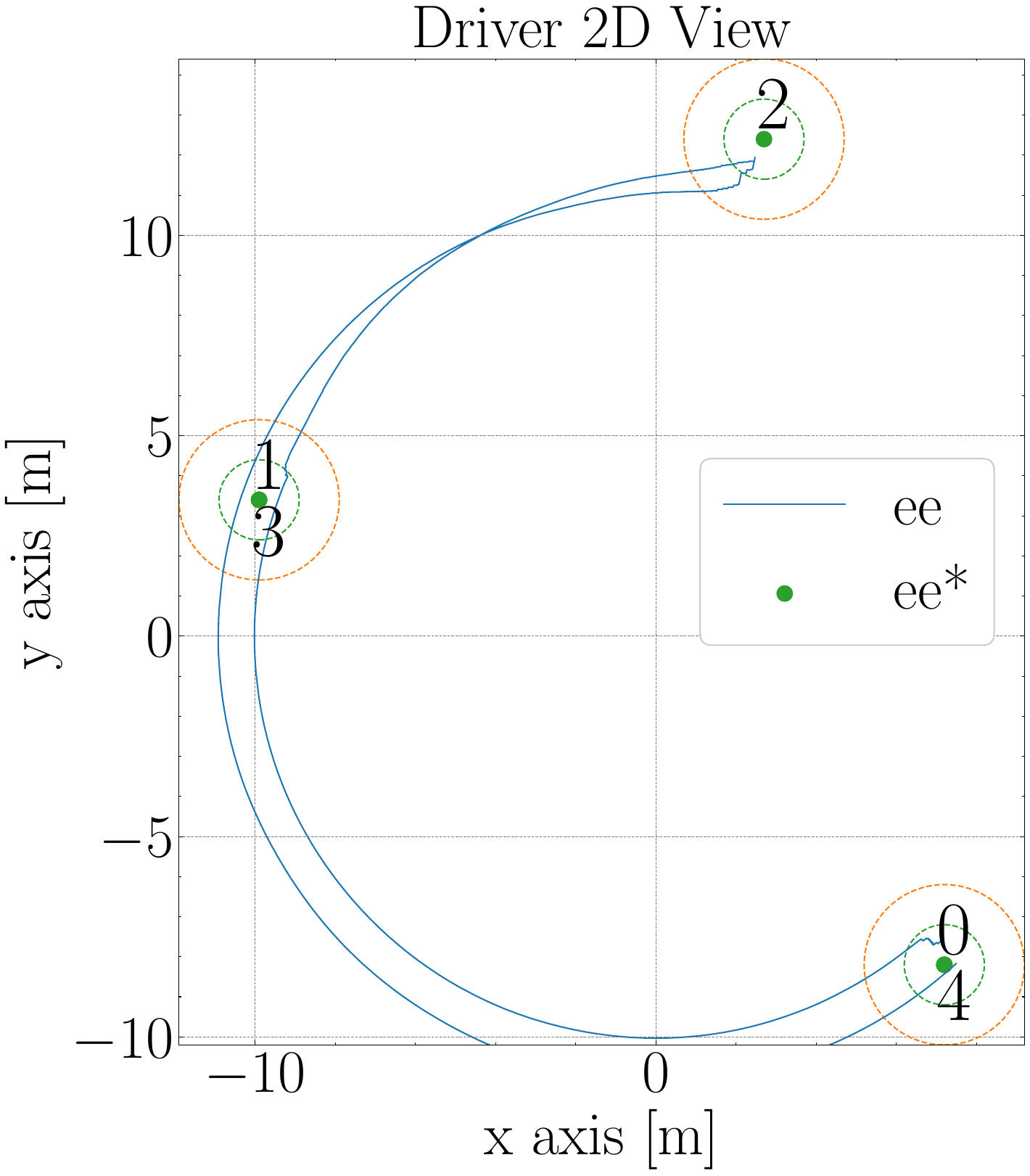}
    \end{minipage}
    \begin{minipage}{0.48\columnwidth}
    \includegraphics[trim={0cm 0cm 0cm 0cm},clip, width=.8\textwidth, left]{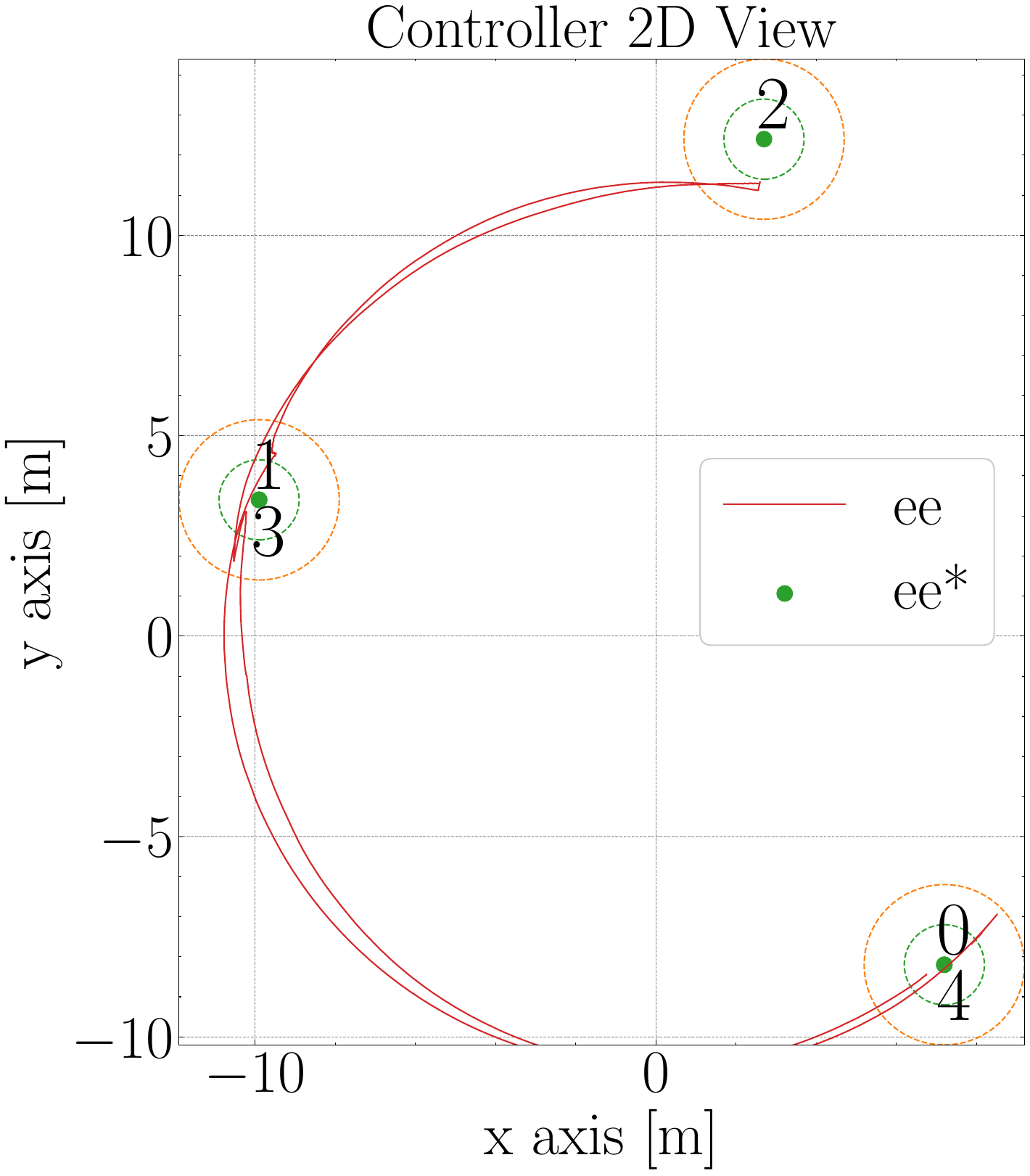}
    \end{minipage}
    \caption{A sequence of four targets at varying heights and distances is provided to both the controller and the driver. While the driver exhibits greater accuracy, the controller utilizes the entire actuation space better. Targets are numbered from 0 (starting position) to 4 (final position).}
    \label{fig:driver_controller}
    \vspace{-0.2cm}
\end{figure}

\begin{figure}
    \centering
    \begin{minipage}{0.45\columnwidth}
    \includegraphics[trim={0cm 0cm 0cm 0cm},clip, width=\textwidth]{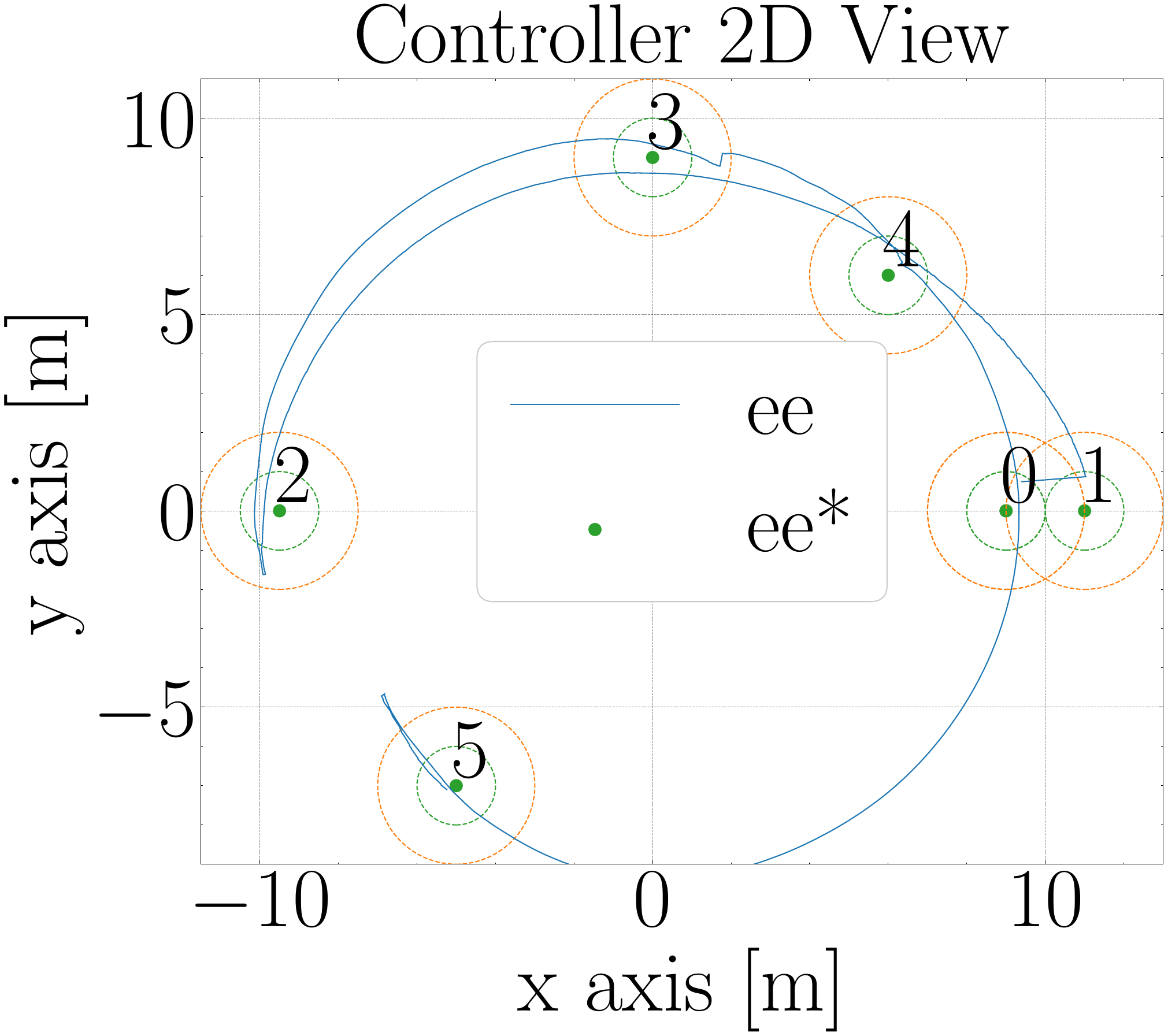}
    \end{minipage}
    \begin{minipage}{0.45\columnwidth}
    \includegraphics[trim={0cm 0cm 0cm 0cm},clip, width=\textwidth]{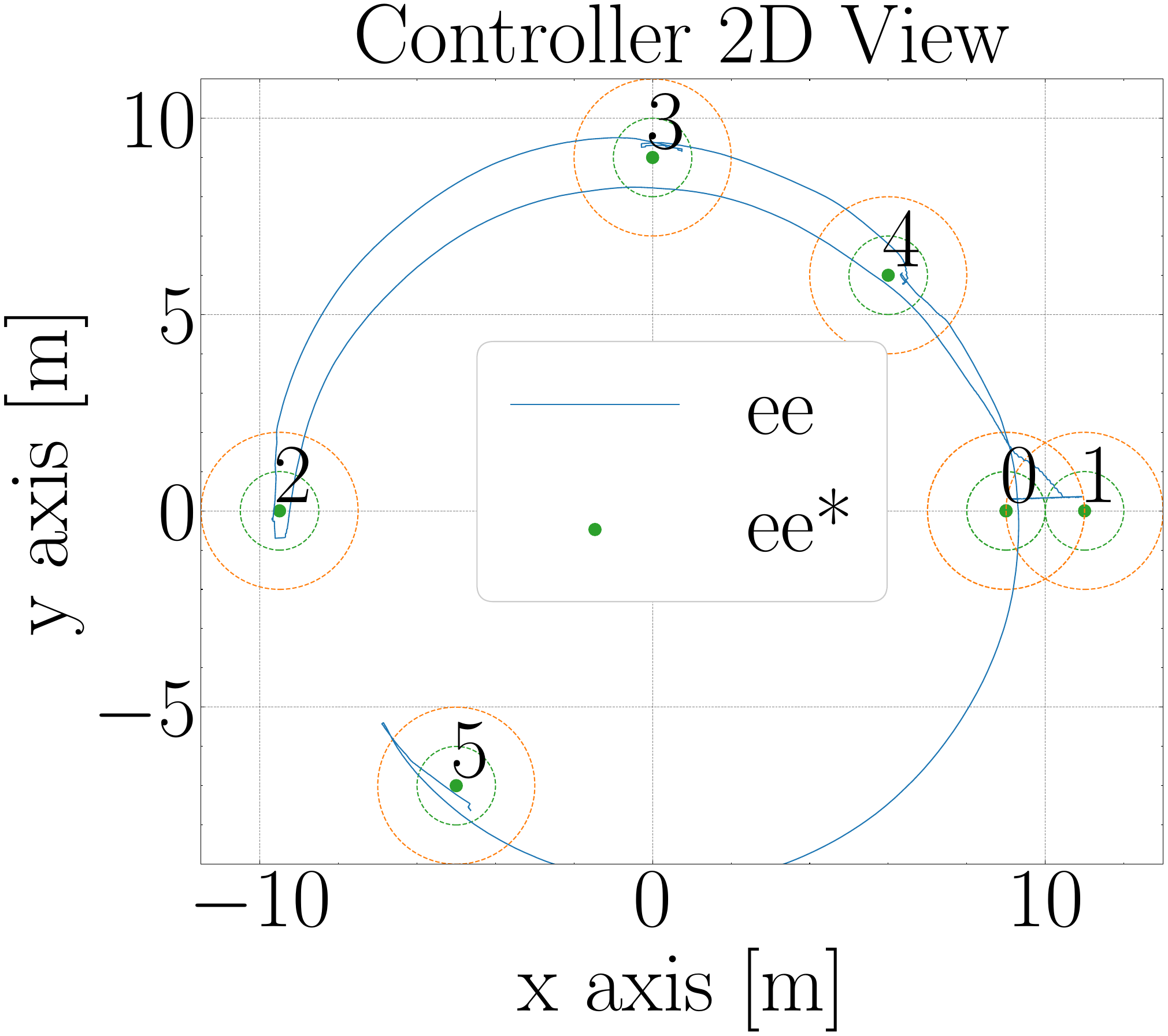}
    \end{minipage}
    \caption{We evaluate five target positions with tolerances of 1 (green circles) and 2~\si{\meter} (orange). On the left is a controller with a 0.5~\si{\second} history without a load in the bucket. On the right, a 0.5~\si{\second} history controller is deployed with an unknown load, performing similarly.}
    \label{fig:controller_2d}
    \vspace{-0.2cm}
\end{figure}

\begin{figure}
    \centering
    \includegraphics[trim={2cm 2.5cm 0.5cm 0.5cm},clip, width=.9\columnwidth]{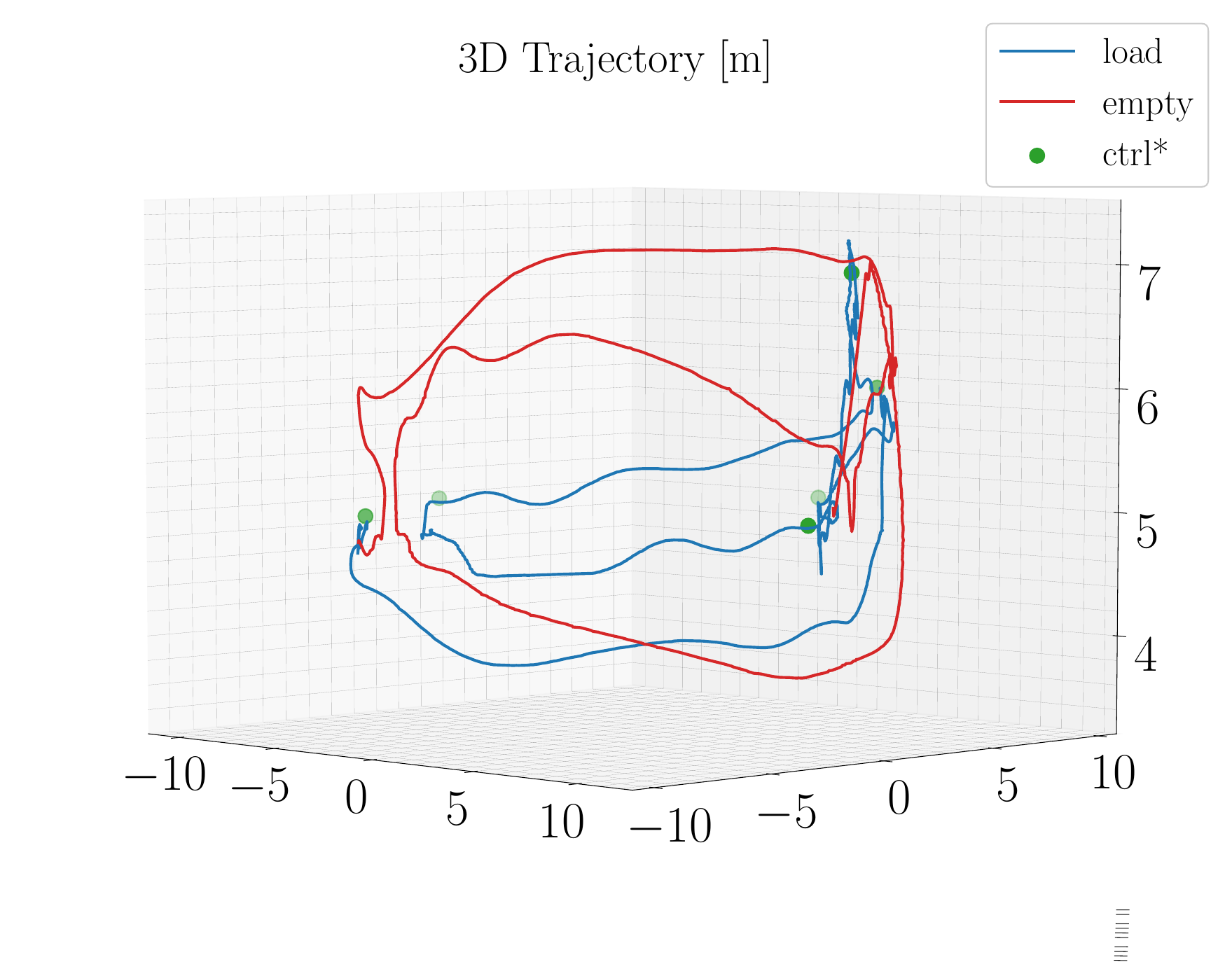}
    \caption{The controller can adopt different strategies according to the state evolution. When carrying a load, a large height variation to reduce tool oscillations is not required because of the increased gripper inertia.}
    \label{fig:controller_space}
    \vspace{-0.2cm}
\end{figure}

\Cref{fig:driver_controller} compares the performance of our controller to an operator with 15 years of experience. Resulting trajectories are similar, but the autonomous controller better exploits the simultaneous actuation of all three joints. As reported in \cref{tab:driver_control}, with comparable slew speeds, the \ac{RL} agent strongly reduces the oscillation of the tool while exhibiting more significant steady-state errors. 
However, a less experienced driver (1/2 years) controls the machine with an error similar to our policy and much larger oscillations. 
In \cref{fig:controller_2d}, we show experiments without access to the load estimation, demonstrating the controller's robustness to variations due to the randomization applied during training. Our approach can handle both empty and full buckets with comparable performance, as reported in \cref{tab:ee_control}. This suggests that it could be extended to the manipulation of heavy objects. 
As seen in \cref{fig:controller_space}, the controller can adjust the final trajectory from observing the dynamics online: with a large load (even if not measured), the gripper is controllable more efficiently, and a direct path can be followed.

\begin{table}[b]
    \vspace{-0.3cm}
    \begin{center}
         \caption{RL Performance \textbf{with} Load Estimation vs. Human Driver. \\ We recorded data from different drivers and controllers and averaged the runs to compute the metrics.}
         \label{tab:driver_control}
         \begin{tabular}{@{}ccccc@{}}
             \toprule
             \textbf{Policy} & \textbf{Speed} [\si{\degree}/\si{\second}] & \textbf{Overshoot} [\si{\degree}] & \textbf{Error} [\si{\meter}] & \textbf{Tool} [\si{\degree}/\si{\second}] \\ \midrule
             Driver & 11.12 & \textbf{0.57} & \textbf{0.593} & 20.23 \\
             Drv. Inex. & \textbf{11.46} & 1.49 & 1.145 & 21.83 \\
             Controller & 11.17 & 7.68 & 1.078 & \textbf{10.49} \\
             \midrule
             Drv. Load & \textbf{11.12} & \textbf{1.55} & \textbf{0.470} & 12.26 \\
             Ctrl. Load & 10.77 & 8.88 & 1.677 & \textbf{6.47} \\ 
             \bottomrule
         \end{tabular}
    \end{center}
    \vspace{-0.5cm}
\end{table}

\renewcommand{\arraystretch}{1.2}
\begin{table}
    \begin{center}
         \caption{RL Performance \textbf{without} Load Estimation. \\ Metrics are obtained by averaging different runs.}
         \label{tab:ee_control}
         \begin{tabular}{@{}ccccc@{}}
             \toprule
             \textbf{Load} & \textbf{Speed} [\si{\degree}/\si{\second}] & \textbf{Overshoot} [\si{\degree}] & \textbf{Error} [\si{\meter}] & \textbf{Tool} [\si{\degree}/\si{\second}] \\ \midrule
             Empty & \textbf{12.43} & 7.79 & 0.843 & \textbf{11.06} \\
             Robust & 11.06 & \textbf{6.02} & \textbf{0.727} & 13.29 \\ \bottomrule
         \end{tabular}
    \end{center}
    \vspace{-0.5cm}
\end{table}

\section{Discussion}

Our experiments reveal that stabilizing the tool requires to superimpose additional motions: our controller implicitly learns to reduce tool oscillations through vertical transitions, using inertia to help damping (\cref{fig:controller_space}). This behavior is enabled by the observation history of 0.5~\si{\second}, which allows to leverage the full machine dynamics. With a shorter observation history, a more conservative and short-sighted control strategy based on tool orientation usually emerges. In practice, these trajectories must be performed accounting for obstacles, a problem which falls beyond the scope of the current work. 

Compared to human operators, one drawback of our approach is the increased overshoot. However, when controlling the slew joint only we were able to achieve competitive results for this metric (\cref{tab:slew_control}). We attribute this problem to the increased task complexity and the additional tool-damping objective, which lead the algorithm to converge to slower and less accurate policies, primarily due to the reward shape: the designed training environment tends to solve the trade-off by prioritizing safety. 
Other factors contributing to the lowered accuracy are the model mismatch arising from the fast-varying inertia, and the limiting assumptions of the \ac{FF} arm controllers.

\section{Conclusions \& Future Work}

In this work, we developed a novel control algorithm to address the automation of material handlers equipped with free-swinging end-effector tools.
For this purpose, we used \ac{RL} to learn a 3D arm position controller, which properly aligns with the task requirements. Our agent controls all \acp{DoF} simultaneously, but with different strategies. The slew hydraulic motor was modeled via \ac{ML} and integrated into the simulation environment, allowing the controller to directly actuate it by learning a correlation between joystick input and velocity output. The arm joints are actuated via simpler velocity controllers, operating over \ac{RL}-provided velocity references.  
Our approach simultaneously handles implicit trajectory planning, grab oscillation, and hydraulic joint control, allowing for a trade-off between tracking accuracy, operational speed, and minimization of the end-effector tool oscillations. 

Our research, a first-of-its-kind control algorithm for large material handling machines, significantly narrows the gap toward deploying autonomous controllers for material handling. 
We demonstrated that \ac{RL} can execute simple tasks competitively compared to average human operators. Despite being less accurate than a very experienced driver, our controller matches their speed and damps the tool under any load conditions more reliably.
The investigation of more powerful architectures, such as \acp{TCN}~\cite{bai2018empirical} and transformers~\cite{vaswani2017attention}, is part of our future research to address the sim-to-real gap. Additionally, we plan to use low-level controllers trained independently of a specific hydraulic model~\cite{Nan2024learning} to improve the velocity tracking performance of the arm joints.  
To tackle collision avoidance with external bodies, we are currently working on incorporating multiple dynamic targets into the tracking objective to develop a path-following tool controller.


\enlargethispage{-2cm}

\bibliographystyle{bibliography/IEEEtran}
\bibliography{bibliography/references}

\end{document}